\documentclass{article}
\usepackage{spconf,amsmath,graphicx}
\usepackage{subfigure}
  
\usepackage{threeparttable}
\usepackage{booktabs}
\title{Cross-Modal Mutual Learning for Cued Speech Recognition}
\name{Lei Liu\textsuperscript{1}, Li Liu\textsuperscript{2,*} \thanks{* Corresponding author: avrillliu@hkust-gz.edu.cn.}}
\address{\textsuperscript{1}Shenzhen Research Institute of Big Data, The Chinese University of Hong Kong, Shenzhen \\\textsuperscript{2}The Hong Kong University of Science and Technology (Guangzhou)}
\begin{document}
%
\maketitle
\begin{abstract}
Automatic Cued Speech Recognition (ACSR) provides an intelligent human-machine interface for visual communications, where the Cued Speech (CS) system utilizes lip movements and hand gestures to code spoken language for hearing-impaired people. Previous ACSR approaches often utilize direct feature concatenation as the main fusion paradigm. However, the asynchronous modalities (\textit{i.e.}, lip, hand shape and hand position) in CS may cause interference for feature concatenation. To address this challenge, we propose a transformer based cross-modal mutual learning framework to prompt multi-modal interaction. Compared with the vanilla self-attention, our model forces modality-specific information of different modalities to pass through a modality-invariant codebook, concatenating linguistic representations with tokens of each modality. Then the shared linguistic knowledge is used to re-synchronize multi-modal sequences. Moreover, we establish a novel large-scale multi-speaker CS dataset for Mandarin Chinese. To our knowledge, this is the first work on ACSR for Mandarin Chinese. Extensive experiments are conducted for different languages (\textit{i.e.}, Chinese, French, and British English). Results demonstrate that our model exhibits superior recognition performance to the state-of-the-art by a large margin.
\end{abstract}
\section{Introduction}
\label{sec:intro}
\begin{figure}[t]
\centering
\includegraphics[width=0.95\linewidth]{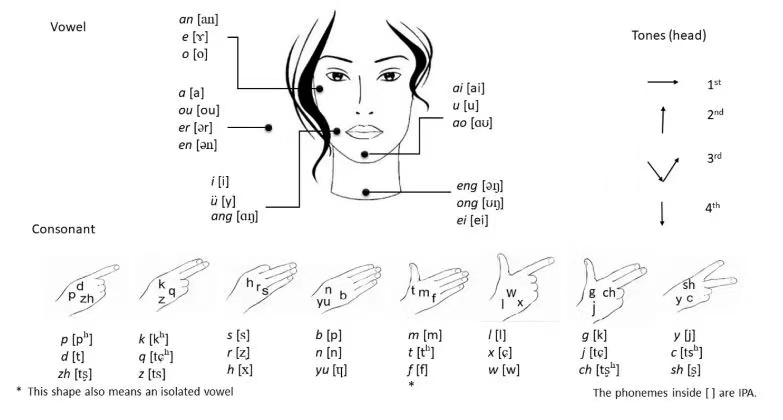}
\caption{The Mandarin Chinese CS system (from \cite{liu2019pilot}).}
\label{chinese}
\end{figure}

Cued Speech (CS) \cite{cornett1967cued,liu2018automatic} is an efficient communication system for hearing impaired people, which leverages lip motions and hand gestures into visual cues. As shown in Figure \ref{chinese} in Mandarin Chinese CS \cite{liu2019pilot,liuobjective}, hand shapes and positions are the supplementary to alleviate the visual ambiguity of similar labial shapes caused by lip reading (\textit{e.g.}, \textit{[p]} and \textit{[b]}). Taking lip and hand as two distinct modalities, Automatic Cued Speech Recognition (ACSR) aims to recognize the multi-modal inputs into linguistic text. The main challenge for ACSR is the natural asynchrony between lip-hand movements, \textit{i.e.}, the hand (\textit{i.e.}, hand shape and hand position) generally move faster than lips to prepare the next phoneme in a CS system, called hand preceding phenomenon \cite{liu2019novel,liu2020re}.

Existing researches mainly assumed lip-hand movements are synchronous by default, thus tended to extract discriminative representations and directly concatenate the high-level features of multi-modal inputs. For example, \cite{heracleous2012continuous,wang2021attention} utilized artificial marks to obtain regions of interests (ROIs) and directly concatenated features of lip and hand. MSHMM \cite{liu2018visual} merged different features by giving weights for different CS modalities. These methods simply exploited feature concatenation as the fusion strategy while ignoring the asynchronous issue, as well as recent knowledge distillation based method \cite{wang2021cross}. To handle asynchronous modalities in the ACSR task, a re-synchronization procedure \cite{liu2020re} was proposed to align visual features using the prior of hand preceding time, which is derived from the statistical information of CS speakers. However, such prior is dependent on speakers and datasets, which is difficult to generalize to different languages. Therefore, there still lacks of effective and flexible fusion strategies for asynchronous CS modalities. Besides, by taking into account the context relationships of phonemes in long-time CS videos, it would be desirable to capture global dependency \cite{vaswani2017attention} over dynamic longer context for solving the above cross-modal fusion task, which could further enhance the interaction of multi-modal sequences in CS. Moreover, previous ACSR task generally utilized phoneme-level annotations on two small scale datasets in French and British English. 

\begin{figure*}[t]
\centering
\includegraphics[width=0.8\linewidth]{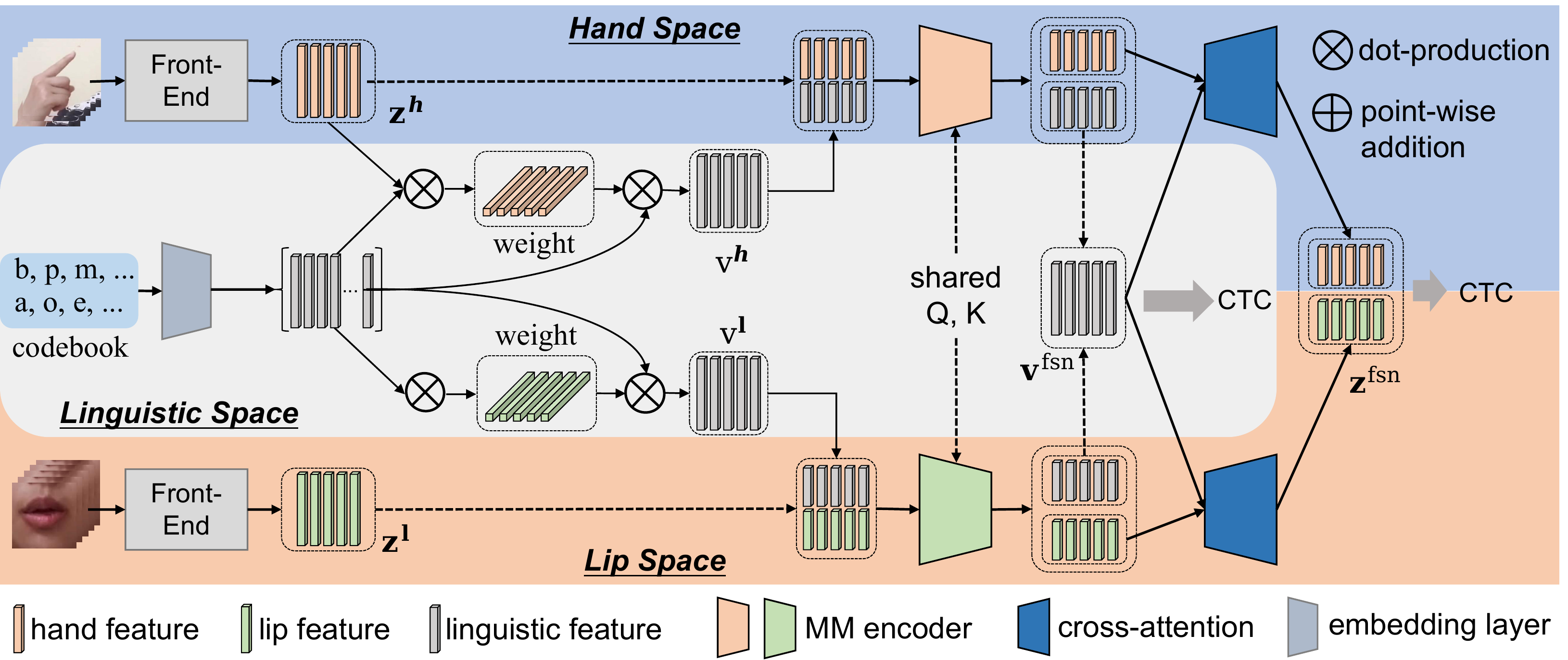}
\caption{The proposed framework for ACSR. A shared frond-end is deployed to extract frame-wise features for each modality and a codebook is to generate modality-invariant linguistic features shared by different modalities. Then a multi-modal (MM) encoder is to capture long-time temporal dependencies, where encoders of different modalities share query and key weights for interactions. Referring to linguistic information, a cross-modal alignment is conducted for visual representations.}
\label{framework}
\end{figure*}

In this study, we firstly establish a novel large-scale multi-speaker CS video dataset for Mandarin Chinese. Then we propose a novel transformer based cross-modal mutual learning framework (see Figure \ref{framework}) to prompt cross-modal information interaction, which achieves feature fusion based on the aligned modalities with the long-time dependencies. In particular, a pre-trained front-end is used to extract frame-wise representations. Then a multi-modal encoder with sharing query and key weights is used to enhance cross-modal interactions. To capture modality-invariant information, our method forces visual representations of different modalities to pass through a linguistic codebook, requiring the model to collate a compact linguistic representation for tokens of each modality. Then linguistic knowledge is exploited to align different modalities based on the cross-attention mechanism.

In summary, the key contributions are: \textbf{(1)} We propose a novel transformer to prompt cross-modal mutual learning for ACSR task based on re-asynchronous modalities with long-time dependencies. Notably, this is the first work that only requires sentence-level annotations to handle asynchronous issue, unlike previous methods relying phoneme-level annotations. \textbf{(2)} We propose to collate a modality-invariant shared representation for multi-modal tokens to obtain linguistic information, which is utilized to guide the alignments for multi-modal data stream. \textbf{(3)} To the best of knowledge, this is the first work on ACSR for Mandarin Chinese. Extensive experiments on three CS benchmarks (\textit{i.e.}, Chinese, French, and British English) demonstrate that our model can significantly outperform the state-of-the-art (SOTA).




\section{Methodology}
\label{method}
\textbf{Problem Formulation.} Given a CS dataset of $N$ quadruples $\mathcal{D}=\{(x_i^l, x_i^g, x_i^p, y_i)\}^N_{i=1}$, corresponding to the lip, hand shape, hand position, and sentence-level label sequences. Our target is to learn linguistic features represented by multi-modal data stream, where lip and hand sequences are complementary to each other as different modalities. Besides, previous works \cite{liu2020re,wang2021cross} often exploited hand region of interests (ROIs) and position coordinates to extract hand shape and position features, respectively. In this work, we experimentally demonstrate that hand ROIs contain both shape and position information, thus only take hand ROIs as the hand input.
The proposed framework is shown in Figure \ref{framework}. 



\subsection{Cross-Modal Mutual Learning}
\noindent \textbf{Front-end.} The front-end adopts a modified ResNet-18 \cite{he2016deep} where the first layer is replaced by a 3D convolutional layer with a kernel size $5 \times 3 \times 3$. The features at the penultimate layer are squeezed along the spatial dimension by global average pooling, resulting in modality-specific features, \textit{i.e.}, $\mathbf{z}^l, \mathbf{z}^g, \mathbf{z}^p \in \mathcal{R}^d$ for lip, hand shape, and hand position, where $d$ is the feature dimension. Element-wise addition operation $\oplus$ is conducted to fuse features of hands via $\mathbf{z}^h=\mathbf{z}^g \oplus \mathbf{z}^p$. To be simplified, we denote $m\in \{l, h\}$ as lip ($l$) and hand ($h$) modalities in the following section, respectively.

\noindent \textbf{Modality-Invariant Linguistic Codebook.} In this work, we aim to extract modality-invariant linguistic representation from the visual features $\mathbf{z}^l$ and $\mathbf{z}^h$. Inspired by \cite{yang2022cross}, we propose to learn a CS codebook to extract linguistic information shared by lip and hand sequences. To this end, an embedding layer is exploited to obtain linguistic codebook basis of cued speech codes. Given the codebook base set $\mathbf{D}=\{b_i\}_{i=1}^{n}$, where $b_i \in \mathcal{R}^d$ and $n$ is the total number of basis. To derive linguistic information, we compute the dot products of the frame-wise visual features ($\mathbf{z}^l$ or $\mathbf{z}^h$) with all bases of $\mathbf{D}$, and apply a softmax function to obtain the normalized weights. This produces the modality-invariant linguistic representations by $\mathbf{v}^m = \text{softmax}(\mathbf{D}\mathbf{z}^m)\mathbf{D}$, where $m \in \{l, h\}$ and linguistic representation is the weighted sum of the bases and the dot product denotes pairwise similarity between $\mathbf{z}^l$ (or $\mathbf{z}^h$) and each basis of $\mathbf{D}$.


\noindent \textbf{Multi-Modal Encoder.}
Given the visual and linguistic representations for lip and hand stream, we aims to force the model to emphasize the temporal information in both modality-invariant and modality-specific flows. Hence, we restrict the cross-modal interactions along with the full sequence to capture long-time dependencies. Formally, we concatenate the visual and linguistic features into a single sequence as:
\begin{equation}
\begin{split}
   \mathbf{u}^{m} = [\mathbf{z}^m||\mathbf{\hat{v}}^m], \quad \text{where} \quad  \mathbf{\hat{v}}^m=g(\mathbf{v}^m, \mathbf{E}_{\text{ling}}^{\text{sub}}) ,
\end{split}
\end{equation}
where $m \in \{l, h\}$ and $[\cdot||\cdot]$ denotes the frame-wise concatenation of the tokens between $\mathbf{z}^l$ (or $\mathbf{z}^h$) and $\mathbf{\hat{v}}^l$ (or $\mathbf{\hat{v}}^h$). We use the projection $\mathbf{E}_{\text{ling}}^{\text{sub}}$ to reduce the dimension of linguistic representations, which can decrease the computation complexity of pairwise attention. The multi-modal encoder stacks a series of vanilla transformer layers, where each modality has its own dedicated value parameters (\textit{i.e.}, $\theta_{v}^{l}$ and $\theta_{v}^{h}$), and shared query $\theta_q$ and key $\theta_k$ parameters. Each lip (hand) token can attend to all other lip (hand) and linguistic tokens via self-attention:
\begin{equation}
    \mathbf{u}^{m}_{L+1} = \text{Transformer}(\mathbf{u}^{m}_{L}; \theta_q, \theta_k,\theta_{v}^{m}),
\end{equation}
where $m \in \{l, h\}$ and $L$ denotes $L$-th transformer layer with vanilla self-attention blocks. This layer allows the information exchanging between visual and linguistic tokens.

We can generalise this model by allowing to linguistic interactions between lip and hand modalities, which is achieved by sharing modality information using compact linguistic vectors. The tokens at layer $L$ are calculated as: 
\begin{equation}
\begin{gathered}
[\mathbf{z}^{m}_{L+1}||\mathbf{\hat{v}}^{m}_{L+1}] = \text{Transformer}(\mathbf{z}^{m}_{L}, \mathbf{\hat{v}}^{\text{fsn}}_{L}; \theta_q, \theta_k,\theta_{v}^{m}),\\
\mathbf{\hat{v}}^{\text{fsn}}_{L}=\frac{1}{2}(\mathbf{\hat{v}}^{l}_{L}+\mathbf{\hat{v}}^{h}_{L}),
\end{gathered}
\end{equation}
where $m \in \{l, h\}$. Since multi-modal attention flows must pass through the shared tight fused linguistics, $\mathbf{z}^{l}$ and $\mathbf{z}^{h}$ can exchange information via $\mathbf{v}^{\text{fsn}}$ within a transformer layer. Thus the model can enhance the cross-modal interactions by using shared linguistic information for the ACSR task.


\noindent \textbf{Visual-Linguistic Alignment (VLA).} To alleviate the naturally asynchronous issue in CS, we aim to integrate aligned information from multiple modalities using cross-attention mechanism. Given the encoded representations of the multi-modal encoder, \textit{i.e.}, $\mathbf{z}^{\text{l}}$,  $\mathbf{z}^{\text{h}}$, and $\mathbf{v}^{\text{fsn}}$, the fused linguistic information $\mathbf{\hat{v}}^{\text{fsn}}$ can be the reference to align lip and hand sequences, since linguistic knowledge is shared by both modalities. We first recover the tight linguistic representation $\mathbf{\hat{v}}^{\text{fsn}}$ into the original dimension by $\mathbf{v}^{\text{fsn}}=g(\mathbf{\hat{v}}^{\text{fsn}}, \mathbf{E}_{\text{ling}}^{\text{up}})$,
where $\mathbf{E}_{\text{ling}}^{\text{up}}$ is a linear projection. Then a cross-attention layer is defined to achieve alignments as:
\begin{equation}
    \mathbf{z}^{m}_{L+1} = \text{Cross-Transformer}( \mathbf{z}^{m}_{L}, \mathbf{v}^{\text{fsn}}; \theta_{q,k,v}).
\end{equation}
Here, the vanilla cross-attention operation \cite{vaswani2017attention} is employed with the shared query, key, and value weights for lip and hand, allowing the multi-modal streams to align with modality-invariant linguistic information. The final fused representation $\mathbf{z}^{\text{fsn}}$ is obtained by the frame-wise concatenation of $\mathbf{z}^{\text{l}}$ and $\mathbf{z}^{\text{h}}$ as $\mathbf{z}^{\text{fsn}}=[\mathbf{z}^{\text{l}}||\mathbf{z}^{\text{h}}]$.

\noindent \textbf{Objective Function.} Connectionist Temporal Classification (CTC) loss \cite{ma2019investigating} shows superior performance for speech recognition tasks. Given the input $\mathbf{x}$ and target sequence $\mathbf{y}$, CTC computes the negative log likelihood of the posterior by summing the probabilities over valid alignment set $\mathcal{A}_{\mathbf{x}, \mathbf{y}}$:
\begin{equation}
-\ln p(\mathbf{y} \mid \mathbf{x})=-\ln \sum_{A \in \mathcal{A}_{\mathbf{z}, \mathbf{y}}} \prod_{t=1}^T p_t\left(a_t \mid \mathbf{x}\right),
\end{equation}
where $T$ is the target length. In this work, we utilize a hybrid CTC architecture to force prediction consistency between visual and linguistic representations:
\begin{equation}
\mathcal{L}_{\text{hyb}} = -\ln p(\mathbf{y} \mid \mathbf{z}^{\text{fsn}}) - \ln p(\mathbf{y} \mid \mathbf{v}^{\text{fsn}}),
\end{equation}
where visual and linguistic representation corresponds to the same target sequence.


\section{Experiments}
\textbf{Datasets.} Three datasets are used to evaluate the performance of the proposed method, including French CS \cite{liu2018automatic}, British English CS \cite{sankar2022multistream}, and the newly collected Mandarin Chinese CS. The details of public CS datasets can refer to Table \ref{dataset}. 

\noindent\textbf{Implementations.} The transformer is randomly initialized while the front-end is pretrained on ImageNet. The multi-modal encoder uses $3$ multi-modal blocks. The mini-batch size is set as 1 \cite{ma2021end}. Following \cite{vaswani2017attention}, the Adam optimizer with $\beta_1 = 0.9$, $\beta_2 = 0.98$ and $\epsilon= 0.05$ is used. The learning rate increases linearly with the first 4000 steps, yielding a peak learning rate and then decreases proportionally to the inverse square root of the step number. The whole network is trained for 50 epochs.

\noindent\textbf{Protocols.} The training and test sentences are randomly split as $4:1$ as shown in Table \ref{dataset}. All methods are evaluated in character error rate (CER) and word error rate (WER) to evaluate the recognition ability on phoneme and word levels.


\begin{table}[!t]
\caption{Details of CS datasets with different languages.}
\begin{center}
\begin{threeparttable}
\resizebox{\linewidth}{!}{
\begin{tabular}{l|c|c|c|c|c}
\hline
Dataset    & French & \multicolumn{2}{c|}{British}	& \multicolumn{2}{c}{Chinese}	\\ \hline
speaker & 1 & 1	& 5	& 1& 4	     \\ \hline
sentence & 238 & 97 & 390	  &1000   & 4000 \\ \hline
character & 12872 & 2741 & 11021	  &32902   & 131581 \\ \hline
word & - & - & -	  & 10562   &  42248 \\ \hline
phoneme & 35 & 44 & 44 & 40 & 40 \\ \hline
gesture & 8 & 8 & 8 & 8 & 8 \\ \hline
position & 5 & 4 & 4 & 5 & 5 \\ \hline
train & \footnotesize 193/10636 & \footnotesize 78/2240 & \footnotesize 312/8924 & \footnotesize 800/26683 & \footnotesize 3200/105372 \\ \hline
test & \footnotesize 45/2236 & \footnotesize 19/501 & \footnotesize 78/2097 & \footnotesize 200/6219 & \footnotesize 800/26209 \\ \hline
\end{tabular}}
\begin{tablenotes}
\item[1] \small \#Train/\#Test is in the form of word/character.
\end{tablenotes}
\end{threeparttable}
\vspace{-0.6cm}
\label{dataset}
\end{center}
\end{table}

\begin{table}[!t]
\caption{Performance comparisons on Chinese CS dataset.}
\begin{center}
\begin{tabular}{c|c|c|c|c}
\hline
Dataset & \multicolumn{4}{c}{Chinese} \\ \hline
\#Speaker                 & \multicolumn{2}{c|}{single} & \multicolumn{2}{c}{multiple}  \\ \hline
Metrics                   & CER   & WER   & CER   & WER  \\ \hline
CNN + LSTM \cite{papadimitriou2021fully}   & 55.4  & 92.8  &  61.4  & 96.1  \\ \hline
CNN + CTC  \cite{he2016deep}         & 35.6  & 78.3  & 41.9  & 83.4     \\ \hline
JLF + COS + CTC \cite{wang2021cross}                & 33.5  & 67.1  & 68.2  & 98.1 \\ \hline
Self-attention \cite{vaswani2017attention}       & 26.1  & 61.8  & 38.8  & 78.6  \\ \hline
Ours                    & \textbf{9.7}  &  \textbf{24.1}   &   \textbf{24.5}    &  \textbf{54.5} \\ \hline
\end{tabular}
\vspace{-0.6cm}
\end{center}
\label{CCS}
\end{table}

\begin{table}[!t]
\caption{Performance comparisons (CER) on British and French CS datasets. WER is unavailable due to lacking of word-level annotations. JLF3 is the previous SOTA.}
\begin{center}
\begin{tabular}{c|c|c|c}
\hline
Dataset & French & \multicolumn{2}{c}{British}  \\ \hline
\#Speaker & single                & single & multiple  \\ \hline
CNN-HMM \cite{liu2018visual}    & 38.0 &- &- \\ \hline
Fully Conv \cite{papadimitriou2021fully}   &  29.2   & 36.3  & - \\ \hline
CNN + LSTM \cite{papadimitriou2021fully} & 33.4 & 43.6 & -  \\ \hline
Transformer \cite{papadimitriou2021fully} & 37.5 & 39.8 & - \\ \hline
Student CE \cite{wang2021cross}    &  35.6   & 47.5  & 37.7 \\ \hline
JLF1 \cite{wang2021cross}         &  27.5   & 38.5  & 34.3 \\ \hline
JLF2 \cite{wang2021cross}          & 27.5   & 36.9  & 31.5   \\ \hline 
JLF3 \cite{wang2021cross}         &  25.8   & 35.1  & 30.3   \\ \hline
Ours               &  \textbf{24.9}   &  \textbf{33.6}  &   \textbf{29.2}  \\ \hline
\end{tabular}
\vspace{-0.6cm}
\end{center}
\label{BCS}
\end{table}

\subsection{Experimental Results}
\textbf{Chinese CS Dataset.} As shown in Table \ref{CCS}, our method achieves SOTA results compared with baseline, \textit{i.e.}, 9.7\% CER in the single-speaker setting and 24.5\% CER in the multi-speaker setting. JLF \cite{wang2021cross} is the previous SOTA method using LSTM, which performs worse than self-attention based approaches. Previous methods still suffer from the asynchronous issue in Chinese CS. Our method can further improve the recognition accuracy via multi-modal alignment. We notice that multi-speaker task is more difficult than single-speaker one due to domain adaption issue.

\noindent\textbf{French\&British CS Dataset.} As shown in Table \ref{BCS}, our method can achieve best results on both French and British CS datasets, outperforming previous SOTA. Compared with LSTM and vanilla transformer, our method benefits from the aligned modalities and can capture effective long-time dependency via multi-modal interaction. The performance improvement is small due to the small data scale of these datasets. The experimental results indicate the effectiveness of the proposed method on small scale datasets.

\begin{figure}[!t]
\centering
\subfigure[Lip Cross-attention ]{\includegraphics[width=.22\textwidth]{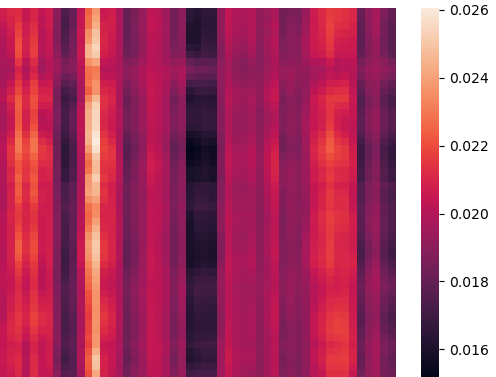}}
\subfigure[Hand Cross-attention]{\includegraphics[width=.22\textwidth]{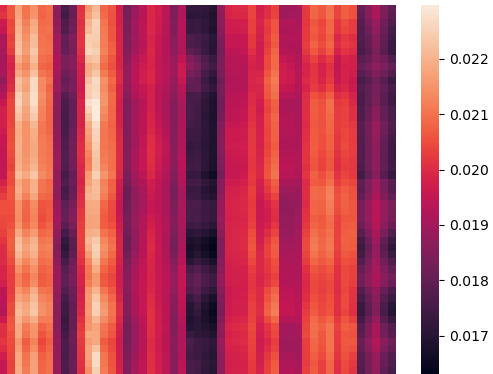}}

\caption{Cross-attention scores on Chinese CS dataset.} 
\label{confusion}  
\end{figure}

\noindent\textbf{Cross-attention Score.} The cross-attention scores of the VLA module are shown in Figure \ref{confusion}. It is observed that lip and hand modalities exhibit similar cross-attention scores, indicating VLA can achieve better cross-modal alignments.

\noindent\textbf{Ablation Studies.} Table \ref{ablation} exhibits the ablation studies for the proposed method. ‘+ self-attn + VLA' only uses linguistic features in the VLA, while lip/hand encoder is without cross-modal interaction. Both codebook and VLA can further improve the performance, outperforming the baselines without codebook and VLA. There exists a significant performance drop when removing each component for our method, indicating the effectiveness of the proposed cross-modal strategy.

\begin{table}[!]
\setlength{\abovecaptionskip}{0cm}
\setlength{\belowcaptionskip}{-0.2cm}
\caption{Ablation Studies. CB is codebook and concat is the concentration of lip and hand features. }
\begin{center}
\begin{tabular}{l|c|c|c|c}
\hline
Dataset & \multicolumn{4}{c}{Chinese} \\ \hline
\#Speaker                 & \multicolumn{2}{c|}{single} & \multicolumn{2}{c}{multiple}  \\ \hline
Metrics (100\%)                   & CER   & WER   & CER   & WER  \\ \hline
CNN + concat                & 35.6  & 78.3  & 41.9  & 83.4 \\ \hline
 + self-attn    & 26.1  & 61.8  & 38.8  & 78.6  \\ \hline
 + self-attn + CB (Eq. 2, 3) & 23.1   & 59.0   & 36.9   & 76.7  \\ \hline
 + self-attn + CB (Eq. 4)   & 16.3   & 43.1  & 34.7   & 72.5  \\ \hline
 + self-attn + VLA & 14.7  & 38.9  & 31.6 &  70.3 \\ \hline
 + self-attn + CB + VLA & \textbf{9.7}  &  \textbf{24.1}   &   \textbf{24.5}    &  \textbf{54.5} \\ \hline
\end{tabular}
\vspace{-0.3cm}
\end{center}
\label{ablation}
\end{table}

\section{Conclusions}
In this work, we proposed a cross-modal mutual learning framework for the ACSR task. We present a multi-modal transformer to capture long-time dependencies for both lip and hand sequences, which transforms modality-specific information into modality-invariant linguistic features via a linguistic codebook. Then modality-invariant linguistic information can guide the cross-modal alignment via the cross-attention operation. The experimental results demonstrate that the proposed approach achieves new SOTA on the ACSR. For the future work, we will be engaged in improving the CS model by decreasing the sentence-level errors.

\textbf{Acknowledgments}. This work is supported by the National Natural Science Foundation of China (No. 62101351), and the GuangDong Basic and Applied Basic Research Foundation (No.2020A1515110376).

\vfill\pagebreak

\bibliographystyle{IEEEbib}
\bibliography{refs}

\end{document}